# Unsupervised Learning to Subphenotype Delirium Patients from Electronic Health Records

Yiqing Zhao
*Department of Preventive Medicine*
Northwestern University
Chicago, USA
yiqing.zhao@northwestern.edu

Yuan Luo (Corresponding Author)
*Department of Preventive Medicine*
Northwestern University
Chicago, USA
yuan.luo@northwestern.edu

*Abstract*—Delirium is a common acute onset brain dysfunction in the emergency setting and is associated with higher mortality. It is difficult to detect and monitor since its presentations and risk factors can be different depending on the underlying medical condition of patients. In our study, we aimed to identify subtypes within the delirium population and build subgroup-specific predictive models to detect delirium using Medical Information Mart for Intensive Care IV (MIMIC-IV) data. We showed that clusters exist within the delirium population. Differences in feature importance were also observed for subgroup-specific predictive models. Our work could recalibrate existing delirium prediction models for each delirium subgroup and improve the precision of delirium detection and monitoring for ICU or emergency department patients who had highly heterogeneous medical conditions.

**Keywords—delirium, subgroup, precision medicine, predictive modeling**

## I. Introduction

Delirium is a common acute onset brain dysfunction in the intensive care unit (ICU) and emergency department (ED) is associated with increased risk of mortality [1, 2], extended hospital stays [3], and long-term cognitive impairment [4]. The Confusion Assessment Method for the ICU (CAM-ICU) [5] and the Intensive Care Delirium Screening Checklist (ICDSC) [6] are the two most widely used tools to assess delirium in an emergency setting. Delirium fluctuates in symptom presentations and severity, making it difficult to diagnose for some subtypes. Traditionally, delirium can be categorized into three subtypes: hyperactive (agitated, restless, hallucinatory and aggressive; prevalence 17%), hypoactive (lethargy, inattentive and slow response to a stimulus; prevalence 10%), and mixed (fluctuating between hyperactive and hypoactive subtypes; prevalence 4%) [7]. Among all three subtypes, hypoactive delirium is the most difficult to detect but has significantly higher mortality [8]. Its presentation includes alteration of consciousness with reduced ability to focus, sustain, or shift attention that cannot be better accounted for by a preexisting, established, or evolving dementia [9]. Difficulties in diagnosing the hypoactive subtype are estimated to contribute to 75% of undetected delirium [10]. Therefore, earlier diagnosis and treatment of delirium, especially hypoactive subtype, in emergency patients are important to improve the outcome of emergency stays, especially when causes of delirium can be managed.

## II. Related Work

Van den Boogaard et al. developed a delirium prediction model – the PREdiction of DELIRium for the Intensive Care patients (PREDELIRIC) model [11] to predict delirium incidences during ICU stays. The model includes 10 risk factors — age, Acute Physiology And Chronic Health Evaluation II (APACHE-II) score, admission group, coma, infection, metabolic acidosis, use of sedatives and morphine, urea concentration, and urgent admission — that can be obtained within 24 hours after ICU admission. Another group utilized machine learning algorithms for predicting ICU delirium and incorporated a much larger feature set, which included patient demographics, admission information, lab measurements, vital signs, comorbidity index, and medication information [12]. The performance of the model is high (F-score > 0.85 for all four algorithms) on their dataset; however, the performance was much lower (F-score < 0.75 for all four algorithms) on external validation dataset Medical Information Mart for Intensive Care IV (MIMIC-III).

As a similar task, Marra et al. proposed an acute brain dysfunction-prediction model (ABD-pm) to predict next-day ABD status [13] among ICU patients. The model was a logistic regression classifier employing 14 known risk factors. A recent work by Yan et al. conducted a multicenter prospective study involving both medical and surgical ICU patients [14] to predict next-day brain function status changes. They trained and externally validated a gradient boosting machine learning model for the previously proposed ABD-pm task and the next-day brain function status changes task. They achieved a greatly improved performance (AUROC = 0.824) verses Marra's benchmark (AUROC = 0.697). In addition to the 14 commonly used risk factors employed in Marra's model, they incorporated 11 questions-based variables extracted from their questionnaire. Through model evaluation, they identified 13 most important features among 25 variables (14 risk factors + 11 questions-based variables).

Recent meta-analysis studies have identified high heterogeneity in the pooled incidence, prevalence, proportion estimates, and risk factors identified for delirium [7, 8], which undermined the generalizability of delirium studies or predictive models. The authors of the meta-analyses speculated that the causes of heterogeneity in the studies might come from the following sources: 1) Lack of a standard set of variables to be incorporated into a multivariable risk factor or outcome model. For example, several studies did not include analgesia or sedative usage in the model. However, sedatives such as benzodiazepines have been identified as a delirium risk factor [15-17]. Therefore, variability in analgesia and sedation practices across settings [18] could contribute to the heterogeneity in meta-analyses. 2) A random-effects meta-regression analysis identified evaluation frequency, study year, study location, the proportion of females, and proportion of mechanically ventilated patients had an independent effect on the heterogeneity of pooled estimates [7]. 3) ICU and ED patient is a heterogeneous population. They often have complex medical conditions and a variety of clinical risk factors (e.g. substance intoxication, or medication side effect) that may contribute to the onset of delirium [19, 20].

Investigating sources of heterogeneity from both clinical and methodologic perspectives can significantly enhance our understanding and guide our practice of delirium prevention, monitoring/diagnosis, and treatment. While such investigation might be difficult, we should at least take into account the existing heterogeneity when building predictive models. In our study, we aimed to build predictive models to predict delirium onset during an emergency stay considering the heterogeneity of delirium from a phenotypic perspective. We took advantage of the new MIMIC-IV database which integrated both ICU and ED visit information. Because delirium may occur outside of ICU stays during an emergency visit and disease burden of delirium was also high among ED patients in general [21], our ICU+ED dataset could provide more insights and capture more comprehensive disease heterogeneity compared to previous studies using only ICU data. Our comprehensive ICU+ED database included crucial data for delirium monitoring and management: patients' labs and medication administration information. We first identified subgroups among delirium patients in the ICU+ED cohort according to their physiological measurements from laboratory tests. We then characterized phenotype features of each delirium subgroup and identified patients with similar features in the entire cohort. Finally, we built one delirium predictive model for each subgroup.

## III. MATERIALS AND METHODS

### A. Data

The data for this study were collected from the MIMIC-IV database [22], an openly available dataset developed by the MIT Lab for Computational Physiology. It contained de-identified electronic health records (EHRs) of 350,000+ patients who had ICD or ED stays at the Beth Israel Deaconess Medical Center (BIDMC, Boston, MA) between 2008 and 2019. For our experiments, we used the most updated version of MIMIC-IV (v0.4), which provides public access to the electronic health record data of 257,366 patients and 524,520 admissions, which includes 69,619 ICU admissions. MIMIC IV not only provided data for patients while they are in an ICU, but also contains information for a patient's entire hospital stay. This allowed us to construct a cohort containing both ICU and ED patients. It also provides medication administration records, with route and dosage information.

### B. Cohort

We retrospectively identified the cohort for this study as patients who had initial delirium onset during an ICU or ED stay (Fig. 1). Delirium onset was determined by: 1) patients first had a delirium ICD code (code list shown in Supplementary Table 1 and was referenced from the Rochester Epidemiology Project codesets [23]) among all admissions; or 2) patients had at least one positive CAM-ICU screening during their stay; or 3) patients were treated with haloperidol, as haloperidol is clinically used only for the treatment of delirium [24]. To align with a previous study [11], we excluded patients who were delirious within 24 hours after admission (determined by CAM-ICU), length of stay (LOS) less than one day, had serious auditory or visual disorders, severe mental retardation, or had serious aphasia (determined by ICD codes). In addition, we excluded patients age <18 from our cohort.

### C. Features

We selected 51 candidate physiological variables characterized in a study by Meng et al. [25]. Additionally, we combined variables measuring the same physiological condition from both labevents and chartevents tables. The 51 candidate physiological variables are listed in Table I. We followed the same data cleaning procedure as in [25]: (1) We converted features with multiple units to their major units. (2) We used the average value for numerical features when duplicate records were found.

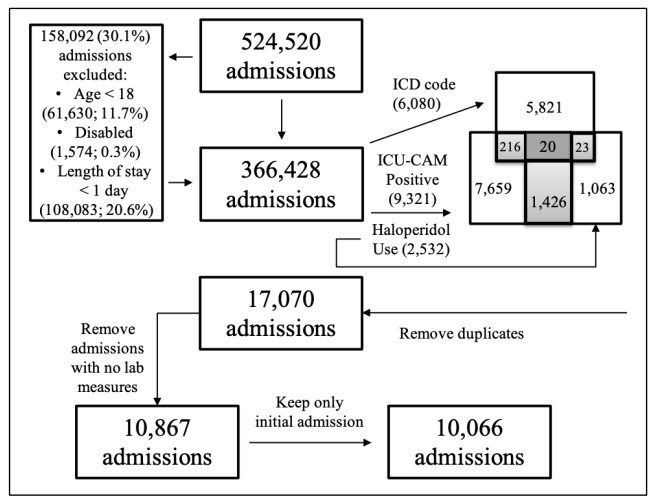

Fig. 1. Delirious admission identification pipeline.

TABLE I. CANDIDATE PHYSIOLOGICAL VARIABLES WITH MEANS AND UNITS

| Candidate physiological variables | Mean (SD) | Unit |
|---|---|---|
| Weight | 80.87 (23.31) | kg |
| Heart Rate | 88.41 (19.98) | bpm |
| Blood Pressure, Systolic | 125.43 (24.77) | mmHg |
| Blood Pressure, Diastolic | 69.17 (17.93) | mmHg |
| Blood Pressure, Mean | 83.13 (20.38) | mmHg |
| Temperature | 36.84 (1.31) | Celsius |
| Blood Albumin | 4.08 (0.39) | g/dL |
| Alkaline Phosphatase | 77.01 (20.16) | IU/L |
| Alanine Transaminase (ALT) | 20.63 (41.99) | IU/L |
| Aspartate Aminotransferase (AST) | 24.35 (36.65) | IU/L |
| Anion Gap | 14.78 (2.62) | mEq/L |
| Base Excess | -1.07 (6.6) | mEq/L |
| Bicarbonate | 25.66 (2.5) | mEq/L |
| Blood Urea Nitrogen (BUN) | 13.77 (3.88) | mg/dL |
| Calcium | 9.09 (0.89) | mg/dL |
| Chloride | 102.33 (3.23) | mEq/L |
| Creatine Kinase/Phosphokinase | 104.46 (163.38) | IU/L |
| Creatinine | 0.85 (0.19) | mg/dL |
| Fraction Of Inspired $O_2$ (FiO$_2$) | 74.36 (26.83) | % |
| Glucose | 113.44 (53.1) | mg/dL |
| Hematocrit | 40.42 (4.29) | % |
| Hemoglobin | 13.74 (1.35) | g/dL |
| International Normalized Ratio | 1.04 (0.24) | - |
| Lactate | 1.4 (0.37) | mmol/L |
| Lactate Dehydrogenase | 193.01 (61.26) | IU/L |
| Magnesium | 2.01 (0.23) | mg/dL |

| Candidate physiological variables | Mean (SD) | Unit |
|---|---|---|
| Mean Corpuscular Hemoglobin | 29.72 (1.4) | pg |
| Mean Corpuscular Hemoglobin Concentration | 33.31 (1.01) | g/dL |
| Mean Corpuscular Volume | 89.9 (4.18) | fL |
| Basophils | 0.17 (0.24) | K/uL |
| Eosinophils | 0.14 (0.16) | K/uL |
| Lymphocytes | 2 (0.76) | K/uL |
| Monocytes | 0.54 (0.34) | K/uL |
| Neutrophils | 4.59 (2.66) | K/uL |
| Partial Pressure Of $O_2$ (PaO$_2$) | 95.09 (9.61) | mmHg |
| Partial Pressure Of $CO_2$ | 40.07 (14.71) | mmHg |
| Positive End-Expiratory Pressure | 5.88 (5.3) | - |
| PH | 7.4 (0.08) | - |
| Platelets | 250.48 (66.37) | K/uL |
| Potassium | 4.16 (0.45) | mEq/L |
| Prothrombin Time | 11.72 (1.38) | sec |
| Partial Prothrombin Time | 29.15 (3.85) | sec |
| Red Blood Cells (RBC) | 4.7 (0.41) | m/uL |
| Red Cell Distribution Width | 13.63 (0.92) | % |
| Saturated $O_2$ (SO$_2$) | 93.93 (11.46) | % |
| Sodium | 138.92 (2.93) | mEq/L |
| Total Bilirubin | 0.55 (0.33) | mg/dL |
| Total $CO_2$ | 25.35 (2.67) | mEq/L |
| Troponin | 0.01 (0.5) | ng/mL |
| Urea Nitrogen | 13.54 (3.76) | mg/dL |
| White Blood Cells | 7.56 (1.8) | K/uL |

### D. Data Pre-processing

(1) Data Filtering: After specifying the list of features, we removed admissions with no measurements for the selected 51 candidate physiological variables and kept only the first delirious admission for each patient in our downstream analysis (see Fig. 1).

(2) Aggregation: For each admission, we averaged abnormal values and normal values separately for each physiological variable. We used averaged abnormal values to represent the physiological variables with abnormal flags recorded for the admission. Otherwise, we used averaged normal values as features for analysis.

(3) Imputation: For cases where value of features were missing, we filled with mean values of corresponding features calculated from normal measurements in the global MIMIC-IV dataset.

(4) We standardized values for each physiological variable into z-scores before downstream clustering tasks.

### E. Clustering for Subgroup Identification

(1) Two clustering methods: We performed clustering analysis on the cohort of 10,066 patients using k-means [26] and hierarchical clustering [27] algorithm. In addition to the 51 physiological features retrieved directly from the database, we calculated six additional clinically meaningful ratios and add them to our feature list.

(2) Optimal number of clusters: We first determined the optimal number of clusters using Silhouette methods [28]. It measured how similar a data point is to other members in its own cluster compared to other clusters. The silhouette width was calculated as in Eq. (1), where a(i) and b(i) represent the average dissimilarity of instance i to all other members in cluster A and B.

$$S(i) = \frac{b(i) - a(i)}{\max(a(i), b(i))} \quad (1)$$

The silhouette width a value within the range [−1, +1], where a high value indicates good matched to the current cluster. If most points in the clusters have a high value, then the clustering result is satisfactory. Otherwise, we may have too many or too few clusters.

(3) Cluster visualization: Finally, we utilized the t-Distributed Stochastic Neighbor Embedding (t-SNE) method [29] to map each patient admission to a two-dimensional space. Two distance metrics (Euclidean distance and cosine similarity) were examined to account for the effect of the metric on the derived clusters.

(4) Cluster validation using Kappa statistics: In order to evaluate the consistency of clustering results, we assessed the agreement of cluster assignments using k-means and hierarchical clustering using Cohen's Kappa Statistics as proposed by Reilly et al. [30]. The rationale behind this evaluation is that if there are no inherent clusters in the data, different methods may find different clusters within data based on their algorithms. On the other hand, if the cluster assignments by a given method are very similar to the ones by another method, we would be confident that the current clustering result is valid.

(5) Feature validation using machine learning classifier: To validate that the selected features were sufficiently representative of the patient subgroups, we trained an XGBoost [31] classifier to re-assign patients to each cluster.

(6) Cluster characterization: For each subgroup, we calculated the means of the physiological variables. Characteristics of subgroups were compared using the chi-square test for categorical variables and Wilcoxon signed rank test for continuous variables. Average of -$\log_{10}$(p-value) was calculated for Wilcoxon signed rank test. We also calculated standard deviations of group means as it was regarded as a better tool than statistical testing to examine heterogeneity between groups when the variance of features was inherently large or when the variance between groups was uneven [32].

### F. Subgroup-specific Delirium Predictive Modeling

(1) Cluster expansion: We used the above trained XGBoost classifier to find non-delirium patients but with similar physiological conditions to each cluster. Same exclusion criteria apply: age <18, disabled, LOS < 1 day, no lab measurements. We also only kept the first admission of each non-delirium patient.

(2) Predictive modeling: Using combined data with delirium and non-delirium patients, we trained a machine learning classifier using logistic regression, random forest, and XGBoost. In addition to the 57 physiological features used in the previous task, we added age, gender, mechanical ventilation status, patients' total admission counts, and admission rank order into our predictive model. Ranks of feature importance were examined to compare classifiers trained on the different subgroups.

## IV. RESULTS

### A. Patient demogrphics

Cohort demographics were summarized in Table II. From a population of 10,066 patients, 59.3% of admissions were in an emergency setting while 43.5% were in an observation setting. There were no obvious gender differences in the cohort. 53.4% of the cohort were seniors (age 65 or over). Overall, in-hospital mortality was 4.6%.

TABLE II. DEMOGRAPHICS AND OUTCOMES SUMMARY OF STUDY COHORT

|  | Number of patients | LOS Admission (day) | Percent Mortality |
|---|---|---|---|
| All patients | 10066 | 12.3 | 4.6% |
| **Sex** | | | |
| Male | 5436 | 12.1 | 4.9% |
| Female | 5431 | 12.5 | 3.7% |
| **Age (years)** | | | |
| 18-24 | 439 | 12.9 | 0.9% |
| 25-44 | 1649 | 13.6 | 1.0% |
| 45-64 | 3404 | 13.2 | 3.3% |
| 65-84 | 3992 | 11.5 | 5.6% |
| 85+ | 1383 | 10.5 | 7.9% |
| **Admission type** | | | |
| EMER | 5965 | 12.6 | 5.8% |
| OBSERVATION | 4376 | 12.1 | 2.7% |
| SURGICAL | 358 | 10.8 | 1.1% |
| ELECTIVE | 168 | 10.1 | 1.8% |

## B. Cluster Assignment and Validation

We determined the optimal number of clusters is four for our dataset (see Supplementary Figure 1). Clustering results both methods to identify the four clusters were plotted in Fig. 2.

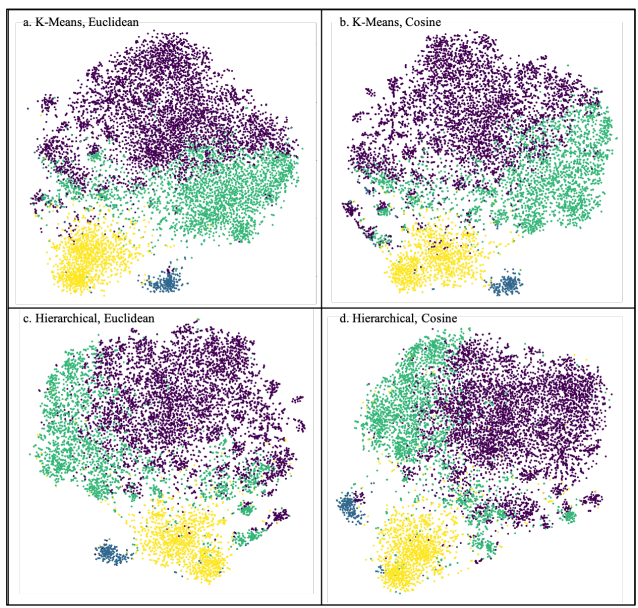

Fig. 2. Clustering reults using (a) k-means, euclidean distance, (b) k-means, cosine similarity, (c) hierarchical clustering, euclidean distance, (d) hierarchical clustering, cosine similarity.

We can see that the sizes of clusters only vary in one major cluster between k-means and hierarchical clustering results. Additionally, distances of four clusters vary only minimally between Euclidean distance plot and cosine similarity plot. Kappa statistics calculated from two clustering results is 0.75, which represent a stable cluster assignment especially for Subgroup 0, 1, and 3. Supplementary Figure 2 showed the agreement percentage between two clustering results in a heatmap.

## C. Subgroup Characteristics

We characterized physiological features and other clinical outcomes for each subgroup using cluster assignments from k-means algorithm (Table III). Standard deviation between subgroup means and average of $-\log_{10}$(p-value) for Wilcoxon signed rank test were listed in Supplementary Table 2. Physiological features with significant between group heterogeneity using the above two measures (value > median for both measures) were highlighted in Table II. Variations of physiological features between subgroups indicates that the identified subgroups were distinctively characterized by 51 physiological variable + 6 ratio variables.

Distinct clinical patterns that indicate underlying medical conditions in each patient subgroup were observed. Subgroup 1 (Fig. 1) has youngest average age but longest LOS. All physiological conditions are the mildest among all four subgroups and are closest to normal. Their in-hospital mortality and emergency admission % are lowest. Subgroup 2 was characterized as diabetic, cardiopathy and coagulopathy, as indicted by a high glucose level, high lactate level, high troponin level and high international normalized ratio (INR), prothrombin time (PT) and partial prothrombin time (PTT). This subgroup has highest mortality rate and oldest age among all four subgroups. Subgroup 3 exhibited patterns of severe renal dysfunction marked by high blood urea nitrogen (BUN), serum creatinine and urea nitrogen, coupled with high inflammation marked by low to decreasing hemoglobin and hematocrit levels. This subgroup also has longest admission history with delirium onset happening relatively late in the disease trajectory. Subgroup 4 was exhibited patterns of severe hepatic and respiratory dysfunction marked by high ventilation rate, low platelet-to-RBC ratio, high aspartate aminotransferase (AST), high lactate dehydrogenase. This subgroup also has shortest admission history with delirium onset happening early in the disease trajectory.

TABLE III. SUBGROUP PHYSIOLOGICAL CHARACTERISTICS (HIGHLIGHTED SIGNIFICANTLY HETEROGENOUS FEATURES IN YELLOW, HIGHEST VALUE IN RED, LOWEST VALUE IN BLUE)

| Candidate physiological variables | Subgroup 1 | Subgroup 2 | Subgroup 3 | Subgroup 4 |
|---|---|---|---|---|
| Patient Count (%) | 5477 (54%) | 261 (3%) | 2980 (30%) | 1348 (13%) |
| Length of stay (days) | 13.2 | 9.6 | 12.3 | 9.1 |
| Age | 58.3 | 69.9 | 66.6 | 65.8 |
| In-hospital Mortality | 1.6% | 13.0% | 6.1% | 12.0% |
| Emergency % | 49.1% | 70.5% | 60.8% | 65.6% |
| Patients' Total Admissions | 5.3 | 4.4 | 7.5 | 4.0 |
| Admission Rank Order | 2.0 | 1.6 | 3.5 | 1.7 |
| Ventilation | 4% | 42% | 9% | 89% |
| Weight | 81.1 | 81.2 | 79.9 | 79.4 |
| Heart Rate | 88.3 | 91.2 | 89.5 | 88.1 |
| Blood Pressure, Systolic | 126.5 | 126.5 | 124.2 | 123.3 |
| Blood Pressure, Diastolic | 70.2 | 70.7 | 68.4 | 67.3 |
| Blood Pressure, Mean | 84.3 | 85.4 | 81.7 | 82.5 |
| Temperature | 36.8 | 36.7 | 36.9 | 37.0 |
| Blood Albumin | 4.1 | 3.5 | 3.7 | 3.6 |
| Alkaline Phosphatase | 84.2 | 91.2 | 101.3 | 85.4 |
| Alanine Transaminase (ALT) | 36.1 | 76.5 | 51.1 | 86.1 |
| Aspartate Aminotransferase (AST) | 43.3 | 110.4 | 79.8 | 129.3 |
| Anion Gap | 15.4 | 16.8 | 15.7 | 15.1 |
| Base Excess | -1.0 | -2.0 | -1.1 | -0.6 |
| Bicarbonate | 25.1 | 23.7 | 24.2 | 24.7 |

| Candidate physiological variables | Subgroup 1 | Subgroup 2 | Subgroup 3 | Subgroup 4 |
|---|---|---|---|---|
| Blood Urea Nitrogen (BUN) | 18.0 | 28.4 | 30.4 | 26.2 |
| Calcium | 9.1 | 8.7 | 8.9 | 8.7 |
| Chloride | 102.1 | 102.6 | 101.5 | 104.3 |
| Creatine Kinase/Phosphokinase | 199.7 | 346.3 | 213.8 | 449.3 |
| Creatinine | 1.0 | 1.5 | 1.7 | 1.3 |
| Fraction Of Inspired $O_2$ (FiO$_2$) | 74.7 | 67.4 | 74.8 | 44.5 |
| Glucose | 127.7 | 169.7 | 139.2 | 152.9 |
| Hematocrit | 39.6 | 33.8 | 30.0 | 31.6 |
| Hemoglobin | 13.3 | 11.2 | 9.7 | 10.5 |
| International Normalized Ratio | 1.2 | 1.7 | 1.4 | 1.5 |
| Lactate | 1.6 | 2.8 | 1.8 | 2.5 |
| Lactate Dehydrogenase | 207.2 | 283.4 | 255.0 | 309.4 |
| Magnesium | 2.0 | 2.0 | 2.0 | 2.0 |
| Mean Corpuscular Hemoglobin | 30.0 | 29.8 | 29.3 | 30.0 |
| Mean Corpuscular Hemoglobin Concentration | 33.5 | 32.9 | 32.2 | 33.0 |
| Mean Corpuscular Volume | 89.9 | 90.9 | 91.1 | 91.3 |
| Basophils | 0.3 | 0.3 | 0.4 | 0.3 |
| Eosinophils | 0.2 | 0.2 | 0.2 | 0.2 |
| Lymphocytes | 1.8 | 1.6 | 1.7 | 1.7 |
| Monocytes | 0.6 | 0.7 | 0.7 | 0.7 |
| Neutrophils | 6.2 | 8.0 | 7.3 | 8.3 |
| Partial Pressure Of $O_2$ (PaO$_2$) | 97.4 | 138.9 | 96.8 | 198.9 |
| Partial Pressure Of $CO_2$ | 40.5 | 41.9 | 41.1 | 42.9 |
| Positive End-Expiratory Pressure | 5.9 | 5.9 | 6.0 | 5.0 |
| PH | 7.4 | 7.4 | 7.4 | 7.4 |
| Platelets | 240.4 | 209.9 | 243.5 | 196.1 |
| Potassium | 4.2 | 4.4 | 4.4 | 4.3 |
| Prothrombin Time | 12.9 | 20.3 | 15.7 | 15.9 |
| Partial Prothrombin Time | 29.8 | 128.8 | 32.2 | 34.1 |
| Red Blood Cells (RBC) | 4.5 | 4.2 | 3.5 | 4.3 |
| Red Cell Distribution Width | 13.7 | 14.4 | 15.8 | 14.1 |
| Saturated $O_2$ (SO$_2$) | 93.6 | 92.6 | 92.8 | 95.4 |
| Sodium | 138.8 | 138.1 | 137.5 | 138.6 |
| Total Bilirubin | 0.7 | 1.2 | 1.0 | 1.1 |
| Total $CO_2$ | 25.4 | 24.3 | 25.4 | 25.5 |
| Troponin | 0.0 | 0.4 | 0.1 | 0.1 |
| Urea Nitrogen | 17.3 | 22.0 | 28.1 | 19.8 |
| White Blood Cells | 9.5 | 12.2 | 10.9 | 13.7 |
| AST:ALT | 1.3 | 1.8 | 1.5 | 1.6 |
| BUN:Creatinine | 18.3 | 22.8 | 21.8 | 23.0 |
| FiO$_2$:PaO$_2$ | 0.8 | 0.7 | 0.9 | 0.4 |
| SO$_2$:FiO$_2$ | 1.3 | 1.5 | 1.2 | 2.2 |
| Neutrophils:Lymphocytes | 5.0 | 8.4 | 8.3 | 8.5 |
| Platelets:RBC | 53.6 | 51.5 | 71.0 | 46.6 |

### D. Feature Set Validation & Predicting Delirium Based on Subgroups

Feature set validation by the XGBoost classifier achieved overall 97.6% accuracy and F-score of 0.977 in the test set with a train-test split ratio of 80:20. This demonstrated that our feature set was sufficient to represent the subgroups.

We applied the classifier on the non-delirium patient population (first admission only) and classified them into the four subgroups that share similar physiological patterns with respect to four delirium subgroups. We ran three classifiers using logistic regression, random forest and XGBoost on the four subgroups as well as an aggregated population (train-test split ratio is 80:20). The resulting patient count and model performance were listed in Table IV. The percentage of delirium patients was lowest in Group 1, which had closest to normal physiological measurement and highest in Group 4, which was highly ventilated with relatively acute conditions. We also observed a slight difference in the predictive modeling performance between groups.

TABLE IV. SUBGROUP COUNTS AND PREDICTIVE MODEL PERFORMANCE

| | All | Subgroup 1 | Subgroup 2 | Subgroup 3 | Subgroup 4 |
|---|---|---|---|---|---|
| Delirium patient | 10066 | 5477 | 261 | 2980 | 1348 |
| Non-delirium patient | 114324 | 66536 | 2717 | 33510 | 11561 |
| Delirium % | 8.1% | 7.6% | 8.8% | 8.2% | 10.4% |
| F score: LR | 0.888 | 0.897 | 0.856 | 0.887 | 0.842 |
| F score: RF | 0.887 | 0.896 | 0.852 | 0.886 | 0.839 |
| F score: XGB | 0.886 | 0.907 | 0.850 | 0.888 | 0.847 |
| AUC: LR | 0.691 | 0.808 | 0.856 | 0.842 | 0.745 |
| AUC: RF | 0.674 | 0.786 | 0.835 | 0.826 | 0.725 |
| AUC: XGB | 0.676 | 0.816 | 0.844 | 0.861 | 0.719 |

In Fig. 3, we plotted feature importance generated through an ensemble of three classifiers: we calculated the mean rank of each feature from three classifiers, with higher value (darker color) representing higher importance. We observed differences in the ranking of features for delirium prediction in different subgroups. We observed that feature importance was quite heterogenous across different groups. Except for 'Patients' Total Admissions', which was quite universally important for all models, 'Age' and 'Ventilation' was most important for the All-patient and Subgroup 1, 3 model. 'Monocytes' was most important for All-patient and Subgroup 2, 4 model. 'Hemoglobin' was most important for Subgroup 1 and 3. 'Blood Urea Nitrogen (BUN)' was most important for Subgroup 2 and 4. Subgroup-specific highly important features included: 'Mean Corpuscular Volume', "Red Blood Cell (RBC)' and 'Urea Nitrogen' for Subgroup 1; 'Heart Rate', 'Blood Pressure', 'Positive End-Expiratory Pressure (PEEP)' and 'Partial Prothrombin Time (PTT)' for Subgroup 2; 'Blood Albumin', 'Base Excess' and 'Mean Corpuscular Hemoglobin Concentration' for Subgroup 3; 'Admission Rank Order' and 'Chloride' for Subgroup 4.

## V. DISCUSSION

Delirium has highly heterogenous presentations and underlying causes. For example, it can be a complication of surgical interventions [33, 34], or it can be caused by inflammation and oxidative stress [35]. Most delirium studies focused on a unique subset of patients e.g., surgery patients [33, 34], elderly patients [35], cancer patients [36]. This has resulted in heterogenous biomarkers being identified for each study, making it difficult to convert evidence to practice [7, 8]. In our work, we studied a heterogeneous population within the emergency setting. Hence, we could better identify different characteristics of delirium patients with different physiological conditions. Current management principles of delirium include two major components: 1) to manage the behavioral symptoms, and 2) to identify and mitigate the underlying risk factors that caused delirious symptoms. Although most risk factors are only minimally modifiable (e.g., cognitive impairment, sleep deprivation, immobility), managing the underlying physiological conditions [37] and monitoring the use of problematic medications [38] can be two important ways for management of delirium [39].

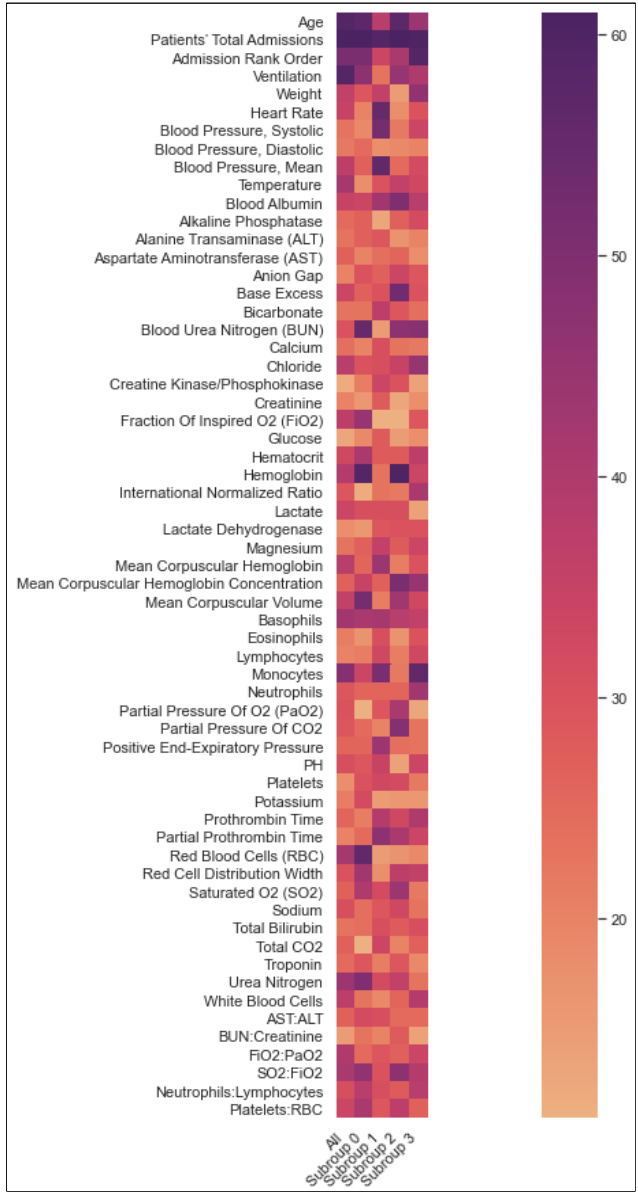

Fig. 3. Average Rank of Feature Importance from Logistic Regression, Random Forest and XGBoost.

Since the underlying physiological conditions can be heterogenous, in our study, we proposed to use clustering algorithms to subgroup delirium population so that we can build subgroup-specific predictive models and identify features with different important for each subgroup. To achieve this, we first compared clustering results from two different clustering methods and confirmed stable clusters exist within the data. Next, we expanded our cohort for predictive modeling using a classifier trained on the delirium cohort to match non-delirium patients to each subgroup. Finally, we trained five predictive models using all eligible patients, and four subgroups. While some commonly predictive features exist e.g. 'Patients' Total Admissions', 'Age', 'Ventilation', 'Hemoglobin', we identified several subgroup-specific features that are important for predicting delirium in each subgroup. Admittedly, missing data is prevalent in clinical dataset and mean imputation may not be the most effective imputation approach. On the other hand, clinical notes often contain complementary information that enable refined patient subphenotyping [40]. We will explore advanced clinical data imputation methods that leverage both cross-sectional and longitudinal correlations [41, 42] in combination with extracted information from clinical notes to improve our subgroup-specific models, better recalibrate existing delirium prediction model for each delirium subgroup, and improve the precision of delirium detection and monitoring for ICU and ED patients who had highly heterogenous medical conditions.

VI. CONCLUSION

In conclusion, our study took into account the heterogeneity in the delirium population and identified four subgroups among the delirium population in our ICU+ED dataset using physiological measurements. Abnormal physiological conditions unique to each subgroup suggested directions for personalized delirium management. We also built four separate predictive models to predict delirium onset during an emergency stay. Unique features more important for one subgroup than the rest indicated potential signals to watch for before delirium onset under different medical conditions.

ACKNOWLEDGMENT

The work is supported in part by NIH grants R01LM013337 and U01TR003528.

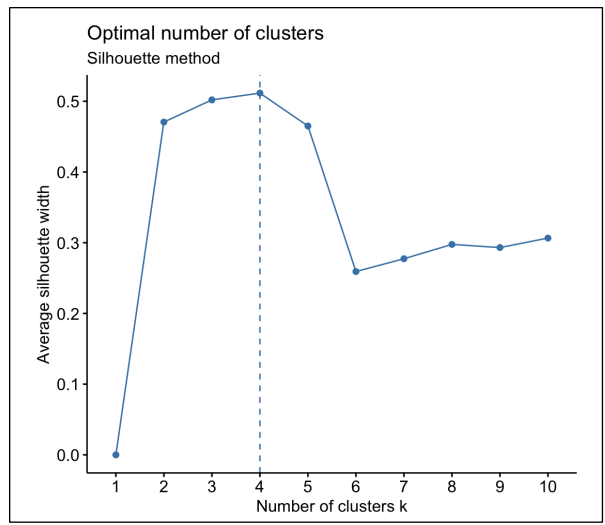

Supplementary Fig. 1. Silhouette width plot to determine optimal number of clusters.

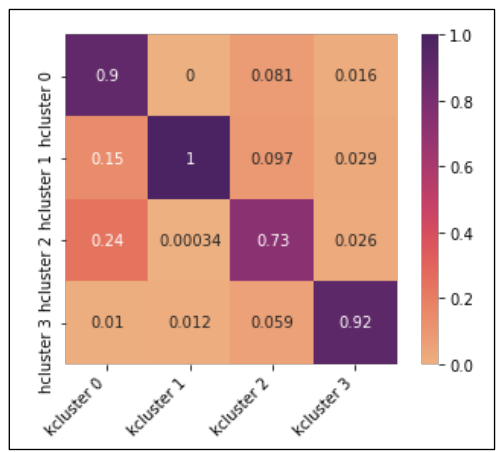

Supplementary Fig. 2. Heatmap of the agreement percentage between two clustering results (kcluster = cluster assignments from k-means algorithm, hcluster = cluster assignments from hierarchical clustering algorithm).

Supplementary Table 1. ICD Codes for Ascertainment of Delirium

| Code | Code Descriptions | Version |
|---|---|---|
| 290.11 | Presenile dementia with delirium | I9 |
| 290.3 | Senile dementia with delirium | I9 |
| 290.41 | Arteriosclerotic dementia with delirium | I9 |
| 291.0 | Alcohol withdrawal delirium | I9 |
| 291.1 | Alcohol - induced persisting amnestic disorder (includes chronic delirium) | I9 |
| 292.81 | Drug-induced delirium | I9 |
| 292.89 | Other specified drug-induced mental disorders (includes chronic delirium) | I9 |
| 293.0 | Acute delirium | I9 |
| 293.1 | Subacute delirium | I9 |
| 308.9 | Unspecified acute reaction to stress (includes exhaustion delirium) | I9 |
| 780.09 | Other alteration of consciousness | I9 |
| F05 | Delirium due to known physiological condition | I10 |
| F01.51 | Vascular dementia with behavioral disturbance | I10 |
| F02.81 | Dementia in other diseases classified elsewhere with behavioral disturbance | I10 |
| F03.91 | Unspecified dementia with behavioral disturbance | I10 |
| F10.221 | Alcohol dependence with intoxication delirium | I10 |
| F10.231 | Alcohol dependence with withdrawal delirium | I10 |
| F10.921 | Alcohol use, unspecified with intoxication delirium | I10 |
| F10.96 | Alcohol use, unspecified with alcohol-induced persisting amnestic disorder | I10 |
| F10.121 | Alcohol abuse with intoxication delirium | I10 |
| F11.121 | Opioid abuse with intoxication delirium | I10 |
| F11.221 | Opioid dependence with intoxication delirium | I10 |
| F11.921 | Opioid use, unspecified with intoxication delirium | I10 |
| F12.121 | Cannabis abuse with intoxication delirium | I10 |
| F12.221 | Cannabis dependence with intoxication delirium | I10 |
| F12.921 | Cannabis use, unspecified with intoxication delirium | I10 |
| F13.121 | Sedative, hypnotic or anxiolytic abuse with intoxication delirium | I10 |
| F13.221 | Sedative, hypnotic or anxiolytic dependence with intoxication delirium | I10 |
| F13.231 | Sedative, hypnotic or anxiolytic dependence with withdrawal delirium | I10 |
| F13.921 | Sedative, hypnotic or anxiolytic use, unspecified with intoxication delirium | I10 |
| F13.931 | Sedative, hypnotic or anxiolytic use, unspecified with withdrawal delirium | I10 |
| F14.121 | Cocaine abuse with intoxication with delirium | I10 |
| F14.221 | Cocaine dependence with intoxication delirium | I10 |
| F14.921 | Cocaine use, unspecified with intoxication delirium | I10 |
| F15.121 | Other stimulant abuse with intoxication delirium | I10 |
| F15.221 | Other stimulant dependence with intoxication delirium | I10 |
| F15.921 | Other stimulant use, unspecified with intoxication delirium | I10 |
| F16.121 | Hallucinogen abuse with intoxication with delirium | I10 |
| F16.221 | Hallucinogen dependence with intoxication with delirium | I10 |
| F16.921 | Hallucinogen use, unspecified with intoxication with delirium | I10 |
| F18.121 | Inhalant abuse with intoxication delirium | I10 |
| F18.221 | Inhalant dependence with intoxication delirium | I10 |
| F18.921 | Inhalant use, unspecified with intoxication with delirium | I10 |

| Code | Description | |
|---|---|---|
| F19.121 | Other psychoactive substance abuse with intoxication delirium | I10 |
| F19.221 | Other psychoactive substance dependence with intoxication delirium | I10 |
| F19.231 | Other psychoactive substance dependence with withdrawal delirium | I10 |
| F19.921 | Other psychoactive substance use, unspecified with intoxication with delirium | I10 |
| F19.931 | Other psychoactive substance use, unspecified with withdrawal delirium | I10 |
| R41.0 | Disorientation, unspecified (delirium not drug/alcohol induced or with dementia or acute/subacute) | I10 |
| F43.0 | Acute stress reaction (exhaustion delirium) | I10 |

Supplementary Table 2. Standard deviation between subgroup means and average of $-\log_{10}$(p-value) for Wilcoxon signed rank test

| Physiological variables | std(mean) | avg($-\log_{10}P$) | Physiological variables | std(mean) | avg($-\log_{10}P$) |
|---|---|---|---|---|---|
| Weight | 0.01 | 6.98 | Basophils | 0.16 | 2.81 |
| Heart Rate | 0.02 | 5.64 | Eosinophils | 0.03 | 0.67 |
| Blood Pressure, Systolic | 0.01 | 10.52 | Lymphocytes | 0.06 | 29.40 |
| Blood Pressure, Diastolic | 0.02 | 13.34 | Monocytes | 0.12 | 7.30 |
| Blood Pressure, Mean | 0.02 | 8.96 | Neutrophils | 0.12 | 10.78 |
| Temperature | 0.00 | 3.68 | Partial Pressure of O2 (PaO2) | 0.36 | 65.89 |
| Blood Albumin | 0.07 | 76.61 | Partial Pressure of CO2 | 0.02 | 0.96 |
| Alkaline Phosphatase | 0.09 | 8.51 | Positive End-Expiratory Pressure | 0.08 | 80.88 |
| Alanine Transaminase (ALT) | 0.37 | 4.77 | PH | 0.00 | 1.88 |
| Aspartate Aminotransferase (AST) | 0.41 | 19.48 | Platelets | 0.10 | 26.60 |
| Anion Gap | 0.05 | 4.99 | Potassium | 0.02 | 4.60 |
| Base Excess | -0.50 | 23.77 | Prothrombin Time | 0.19 | 76.86 |
| Bicarbonate | 0.02 | 7.89 | Partial Prothrombin Time | 0.86 | 88.74 |
| Blood Urea Nitrogen (BUN) | 0.21 | 52.28 | Red Blood Cells (RBC) | 0.10 | 45.21 |
| Calcium | 0.02 | 13.13 | Red Cell Distribution Width | 0.06 | 31.22 |
| Chloride | 0.01 | 15.23 | Saturated O2 (SO2) | 0.01 | 58.22 |
| Creatine Kinase/Phosphokinase | 0.39 | 3.18 | Sodium | 0.00 | 7.75 |
| Creatinine | 0.20 | 22.50 | Total Bilirubin | 0.24 | 2.35 |
| Fraction Of Inspired O2 (FiO2) | 0.22 | 27.56 | Total CO2 | 0.02 | 3.20 |
| Glucose | 0.12 | 37.51 | Troponin | 1.09 | 6.22 |
| Hematocrit | 0.12 | 17.16 | Urea Nitrogen | 0.21 | 35.00 |
| Hemoglobin | 0.14 | 22.33 | White Blood Cells | 0.16 | 44.94 |
| International Normalized Ratio | 0.16 | 88.95 | AST:ALT | 0.16 | 14.65 |
| Lactate | 0.25 | 36.03 | BUN:Creatinine | 0.10 | 14.30 |
| Lactate Dehydrogenase | 0.17 | 44.89 | FiO2:PaO2 | 0.34 | 16.18 |
| Magnesium | 0.00 | 1.69 | SO2:FiO2 | 0.30 | 21.87 |
| Mean Corpuscular Hemoglobin | 0.01 | 7.18 | Neutrophils:Lymphocytes | 0.23 | 22.99 |
| Mean Corpuscular Hemoglobin Concentration | 0.02 | 60.66 | Platelets:RBC | 0.19 | 43.18 |
| Mean Corpuscular Volume | 0.01 | 4.86 | | | |